\pdfoutput=1

\documentclass[11pt]{article}
\RequirePackage[dvipsnames,x11names]{xcolor}

\usepackage[]{ACL2023}

\usepackage{times}
\usepackage{latexsym}
\RequirePackage{amsmath,amssymb,bm}

\usepackage[T1]{fontenc}

\usepackage[utf8]{inputenc}

\usepackage{microtype}
\usepackage{booktabs}
\makeatletter
\def\paragraph{\@startsection{paragraph}{4}{\z@}{.5ex plus
  0.1ex minus .2ex}{-0.5em}{\normalsize\bf}}
\g@addto@macro{\normalsize}{%
\setlength{\abovedisplayskip}{2pt plus1pt}%
\setlength{\abovedisplayshortskip}{2pt plus1pt}%
\setlength{\belowdisplayskip}{2pt plus1pt}%
\setlength{\belowdisplayshortskip}{2pt plus1pt}}

\textfloatsep \intextsep
\floatsep \intextsep
\makeatother

\RequirePackage[skip=.3ex plus.2ex]{caption}

\author{Vishwajeet Kumar$^*$ \\ IBM Research \\ India \\ 
        \And  Yash Gupta$^*$ \\ IIT Bombay \\ India \\ \And Saneem Chemmengath \\ IBM Research$\dagger$ \\ India \\
        \AND Jaydeep Sen \\ IBM Research \\ India \\
        \And Soumen Chakrabarti \\ IIT Bombay \\ India \\
        \And Samarth Bharadwaj  \\ IBM Research$\dagger$ \\ India \\
        \And Feifei Pan \\ IBM Research$\dagger$ \\ US \\ }

\def\shortname{MITQA}

\def\ztitle{Multi-Row, Multi-Span Distant Supervision For Table+Text Question Answering}

\title{\ztitle}

\RequirePackage[inline,shortlabels]{enumitem} \setlist{nosep}
\RequirePackage{adjustbox}
\usepackage[noEnd]{algpseudocodex}

\usepackage{algorithm}
\RequirePackage{soul}
\RequirePackage{colortbl}

\RequirePackage[textsize=scriptsize,color=yellow!20,
linecolor=orange,bordercolor=orange]{todonotes}
\usepackage{amsmath}

\DeclareMathOperator*{\argmax}{argmax}

\newcommand{\eat}[1]{} 

\begin{document}
\maketitle

\begin{abstract}
Question answering (QA) over tables and linked text, also called TextTableQA, has witnessed significant research in recent years, as tables are often found embedded in documents along with related text. HybridQA and OTT-QA are the two best-known TextTableQA datasets, with questions that are best answered by combining information from both table cells and linked text passages. A common challenge in both datasets, and TextTableQA in general, is that the training instances include just the question and answer, where the gold answer may match not only multiple table cells across table rows but also multiple text spans within the scope of a table row and its associated text.  This leads to a noisy multi-instance training regime. We present \shortname, a transformer-based TextTableQA system that is explicitly designed to cope with distant supervision along both these axes, through a multi-instance loss objective, together with careful curriculum design.  Our experiments show that the proposed multi-instance distant supervision approach helps \shortname{} get state-of-the-art results  beating the existing baselines for both HybridQA and OTT-QA, putting \shortname{} at the top of HybridQA leaderboard with best EM and F1 scores on a held out test set.

\end{abstract}
\section{Introduction}
\label{sec:Intro}
\def\thefootnote{*}\footnotetext{Equal Contribution}\def\thefootnote{\arabic{footnote}}
\def\thefootnote{$\dagger$}\footnotetext{Work done while at IBM Research}\def\thefootnote{\arabic{footnote}}

\begin{figure*}
  \includegraphics[width=\hsize]{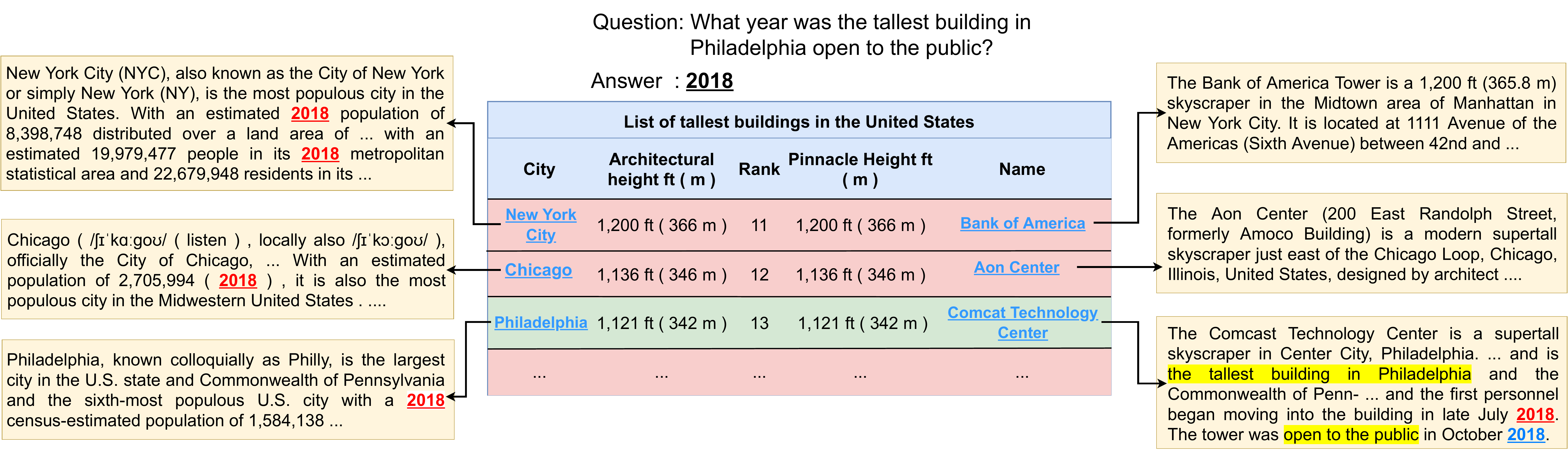}
  \caption{An instance of question answering over hybrid context of table and text (from HybridQA).  Gold answer in correct context is highlighted in blue and gold answer appearing in irrelevant context is highlighted in red. The context used to arrive at the answer in the correct passage is shaded in yellow. The relevant row to be retrieved is shaded green and irrelevant rows are shaded red.}
  \label{fig:motivating_example}
\end{figure*}

Transformer-based question answering (QA) methods have evolved rapidly in recent years to handle open-domain, multi-hop reasoning over retrieved context paragraphs.
Many existing QA datasets and benchmarks measure performance over homogeneous data sources, such as text \cite{DBLP:journals/corr/RajpurkarZLL16,DBLP:journals/corr/ChenFWB17,DBLP:journals/corr/JoshiCWZ17,DBLP:journals/corr/abs-1903-00161} and more recently tables~\cite{pasupat2015compositional, zhong2017seq2sql,nsm,herzig2020tapas, yin2020tabert}. Even though real-world documents often contain tables embedded in free form text, QA over such a hybrid corpus, i.e., a combination of tables and text --- a.k.a.\ \textbf{TextTableQA} --- remains relatively unexplored.
As illustrated in \figurename~\ref{fig:motivating_example}, even a relatively simple table from Wikipedia often references several entities, definitions or descriptions from the table elements.
A question may be best answered by matching some parts of it to table elements and other parts to linked text spans.  Existing Transformer-based QA solutions need significant modifications to score such heterogeneous corpus units.
A key challenge is to reduce the cognitive burden of supervision to (question, answer) pairs, without humans having to identify the specific table cell or text span where the answer was mentioned.
In TextTableQA, such `distant' supervision is particularly challenging because it occurs along two distinct axes:
\begin{enumerate*}[(1)]
\item There could be multiple rows and associated passages that mention the answer string; and
\item Even for a specific table row with linked passages, the same candidate answer may occur in multiple text spans.
\end{enumerate*}
Many of them may be spurious and detrimental to system training.

In response, we present \textbf{\shortname} --- a TextTableQA system specifically engineered to address the above challenges.
%
\shortname{} defines each table row, together with linked passages, as the fundamental \emph{retrieval unit}.
To adapt to memory-hungry Transformer networks, constrained by the number of input tokens they can efficiently process, 
\shortname{} uses a novel query-informed passage filter to prepare a contextual representation of each retrieval unit.
\shortname{} then uses an early interaction (cross attention) Transformer network to score retrieval units.
While training \shortname, its most salient features are multi-instance loss functions and data engineering curricula to tackle distant supervision, along both the multi-row and multi-span axes.
Many of the above challenges are not faced by homogeneous text-only or table-only QA systems.
We report results on extensive experiments on two recent TextTableQA challenge data sets, HybridQA and OTT-QA, where our system outperforms baselines and is currently at the top of HybridQA\footnote{\protect\url{https://competitions.codalab.org/competitions/24420\#results}} leaderboard.
Source code is available at \url{https://github.com/primeqa/primeqa}.

\section{Related Work}

TableQA has gained much popularity in recent years, resulting in diverse approaches including semantic parsing-based \citep{pasupat2015compositional, zhong2017seq2sql, nsm, krishnamurthy-etal-2017-neural, dasigi-etal-2019-iterative} and more recently BERT-based \citep{bert} systems for table encoding by, inter alia, \citet{herzig2020tapas, yin2020tabert, RCI}. A more realistic application scenario is ``TextTableQA'' where tables are often embedded in documents and a natural language query needs to combine information from a table as well as its correlated textual context to find an answer. 

HybridQA \citep{hybridqachen2020} pioneered a TextTableQA benchmark, with Wikipedia tables linked to relevant free-form text passages (e.g., Wikipedia entity definition pages).  They curated questions which need information from both tables and text to answer correctly.   They also proposed HYBRIDER as the first system in TextTableQA with an F1 score of 50\%, leaving much scope for improvement.  The OTT-QA \citep{ottqahen2021open} benchmark extended HybridQA to an open domain setting where a system needs to retrieve a relevant set of tables and passages first before trying to answer a question. Moreover, the links from table and passage are not provided explicitly.  To our knowledge, no existing TextTableQA system \cite{hybridqachen2020,ottqahen2021open,zhong2022reasoningcarp} attempts to handle the challenge of multiple candidate instances arising from distant supervision during system training, owing to multiple matching table rows and multiple matching spans within a row and its linked text.  Our experiments with HybridQA and OTT-QA show that superior handling of multi-instance matches by \shortname{} improves QA accuracy.

\section{Preliminaries}

\def\meta#1{\ensuremath{#1\text{\sffamily .meta}}}
\def\cols#1{\ensuremath{#1\text{\sffamily .cols}}}
\def\rows#1{\ensuremath{#1\text{\sffamily .rows}}}
\def\chdr#1{\ensuremath{#1\text{\sffamily .hdr}}}
\def\pass#1{\ensuremath{#1\text{\sffamily .psg}}}
\def\spans{\ensuremath{\text{\sffamily spans}}}

\subsection{Notation}

$T$ denotes a set of tables, each table being denoted as~$t$.  
Title, caption, and other available metadata of table $t$ is accessed as $\meta{t}$.
Table $t$ has $\rows{t}$ rows and $\cols{t}$ columns.  Its column headers are denoted $\chdr{t}$.  (Row headers may also assume a similar salient role, but we limit notation to column headers for simplicity of exposition.)  $[N]$ denotes the set of indices $\{1,\ldots,N\}$.  For $r\in[\rows{t}]$, the $r$th row is denoted $t[r,\star]$.  For $c\in[\cols{t}]$, the cell at position $(r,c)$ is written as $t[r,c]$.  The $c$th column header cell is denoted $\chdr{t}[c]$.  The set of passages linked with the row $r$ of table $t$ is denoted by $\pass{t[r,\star]}$.  A passage $p$ is a sequence of tokens. The set of all token spans in $p$ is denoted by $\spans(p)$.  One token span is denoted $\sigma \in \spans(p)$.  A set of such spans is denoted~$\Sigma$.

\begin{figure*}
\centering
\includegraphics[width=\hsize]{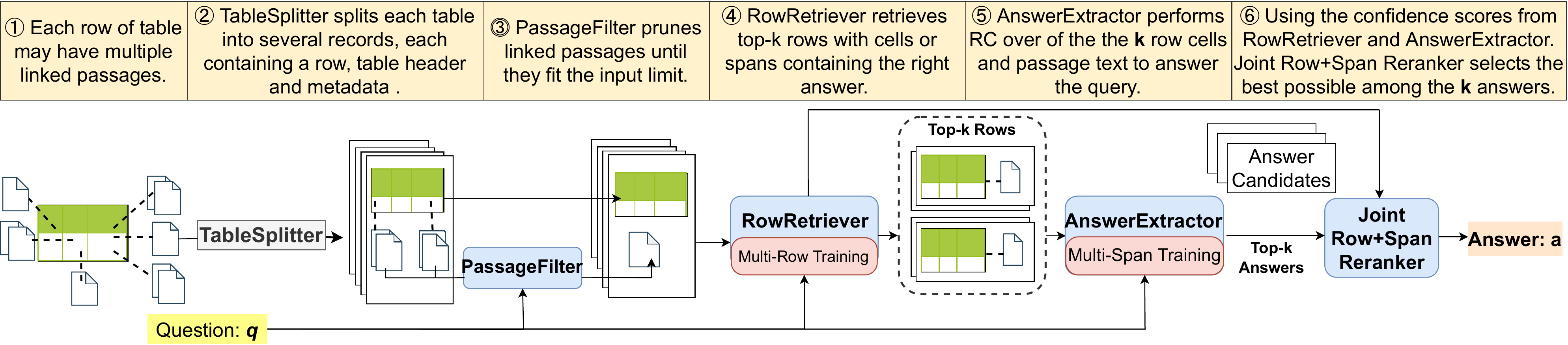}
\caption{\shortname{} system sketch.  TableRetriever and RowPassageLinker are not shown.}
\label{fig:arch}
\end{figure*}

\subsection{Task Definition}

Given a question $q$ (modeled as a sequence of tokens) and a table $t$ together with linked text, the task is to find a
relevant row $r$, and then an answer text $a$, which can be a cell from
$t[r,\star]$, or a span from $\spans(\pass{t[r,\star]})$.
In HybridQA, the table $t$ and associated linked passages are provided along with the question~$q$.  In contrast, for OTT-QA, the correct table $t$ and linked passages need to retrieved from a corpus of tables and initially unconnected passages---a more challenging setting.

\subsection{System Overview}

Figure~\ref{fig:arch} shows the overall architecture of \shortname. 
In some workloads (e.g., HybridQA), a question comes already associated with a table and its linked text.  In other ``open domain'' workloads (e.g., OTT-QA), tables and linked passages must be retrieved from a large corpus by a \textbf{TableRetriever}.
The \textbf{TableSplitter} segments the table $t$ into \emph{retrieval units}, each comprising one row $r$ (i.e., all cells in $t[r,\star]$) and its linked passages $\pass{t[r,\star]}$.
For data sets (like OTTQA) which are not provided pre-linked, the \textbf{RowPassageLinker} module links spans in table cells to corpus passages to prepare the retrieval units.
To score retrieval units, we will use an early interaction (cross-attention) Transformer network, to which we will feed the question and a retrieval unit, suitably encoded into text.
Rather than naive truncation, or expensive hierarchical encodings, we use a question-sensitive \textbf{PassageFilter} to select a subset of passages $\operatorname{PassageFilter}(t,r,q) \subseteq \pass{t[r,\star]}$ to retain with each candidate row.
The \textbf{RowRetriever} can then identify the most relevant retrieval units.
Next, an \textbf{AnswerExtractor} module selects the answer span as a cell from $t[r,\star]$ or as a token span from a passage $p \in \pass{t[r,\star]}$ linked to the row $t[r,\star]$.

Distant supervision (as described above), and the consequent need for multi-instance learning, are handled by three modules: RowRetriever, AnswerExtractor, and a final 
\textbf{RowSpanReranker}.  RowRetriever employs a special loss function that can handle spurious matches of the gold answer in multiple rows and associated retrieval units \citep{mil_97, mi_svm}.  AnswerExtractor employs a data programming \citep{Ratner2016DataProg} curriculum to a similar end.  
The final Reranker module refines the score for each answer candidate, based on a learned weighted combination of RowRetriever and AnswerExtractor confidence scores. We describe the most important components of \shortname{} in Section~\ref{sec:mitqa_system} and defer the rest to Appendix\,\ref{app:mitqa_arch_details}.

\section{\shortname{} System Architecture}
\label{sec:mitqa_system}


In this section we first describe the modules shown in Figure\,\ref{fig:arch}, that are shared for closed-domain (table and linked text provided, as in HybridQA) and open-domain (OTT-QA) applications.  After that we describe TableRetriever and RowPassageLinker that are needed for open-domain scenarios.

\subsection{PassageFilter}
\label{subsec:PassageFilter}

The total tokens in passages linked to a row can be large, exceeding the input capacity of BERT-like models.  Efforts \citep{Beltagy2020Longformer, Zaheer2020BigBird} have recently been made to remove these capacity limits, but at the cost of additional complexity, unsuited for our fine-grained application to table rows.  In any case, the query has a critical role in determining the utility of each passage linked to a row.  Our PassageFilter module orders the linked passages such that the prefix that fits within the input capacity of a BERT-like model is likely to be the most valuable for judging the relevance of a row.  More details are in Appendix\,\ref{app:PassageFilter}.


\subsection{RowRetriever}
\label{subsec:row_retriever}

Given question $q$ and table $t$, the task of RowRetriever is to identify the correct row $r$ from which the answer can be obtained, either as a cell $t[r,c]$ from the $c$th column, or a span from a passage in $\pass{t[r,\star]}$.
We implement RowRetriever by training a BERT-based sequence classification model \citep{bert} on a binary classification task with correct rows to be labelled as $1$s and the rest as~$0$s.  Suppose the columns of $t$ are indexed left-to-right using index~$c$. Then $\chdr{t}[c]$ and $t[r,c]$ are the header and cell in column $c$.  The input $\bm{x}$ to the BERT encoder is fashioned as:
\begin{multline} 
\texttt{[CLS]} \, q  \, \texttt{[SEP]} \hspace{-1em}
\operatorname*{\Big\Vert}_{c \in [\cols{t}]} \hspace{-1em}
\colorbox{gray!10}{$\chdr{t}[c]\;\text{is}\;t[r,c]\,\texttt{[DOT]}$} \\ 
\texttt{[SEP]} \, \meta{t} \, \texttt{[DOT]} \hspace{-2em}
\operatorname*{\Big\Vert}_{p \in \operatorname{PassageFilter}(t,r,q)} \hspace{-3em}
\colorbox{gray!10}{$p \, \texttt{[DOT]}$}
\label{eqn:rr_input_format}
\end{multline}
where `$\Vert$' is the concatenation operator and \text{`is'} is literally the word `is'. \texttt{[DOT]} and \texttt{[SEP]} are separator 
tokens.  In words, we concatenate:
\begin{enumerate*}[(1)]
\item the question~$q$;
\item phrases of the form ``header is cell-value'', over all columns;
\item table metadata (title etc.); and
\item passages linked to the given row, that survive through PassageFilter;
\end{enumerate*}
before passing into a BERT-Large encoder in a specific format to get a suitable latent states.  The \texttt{[CLS]} embedding output by BERT is sent to a feed-forward neural network to make the label prediction. During inference, all \{question, row\} pairs are passed through this sequence classifier. The row with the largest score for class 1 is identified as the chosen row.

\paragraph*{Distant supervision of RowRetriever:}
A row retrieval system that expects supervision in the form of gold rows exacts a high cognitive burden from annotator in preparing training instances. In the case of HybridQA and OTT-QA, we only have final answer-text as supervision, not relevant row/s, cell/s or text span/s. Given a table with connected passages and a question, we identify potential gold rows by exact string matching answer-text on rows (cells and linked texts).
\begin{figure}[t]
\centering
\includegraphics[trim=0 0.5mm 0 0.5mm, clip, width=0.6\hsize]{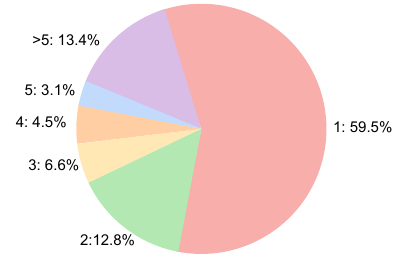}
\caption{Distribution of number of rows containing the answer-text in the training set of HybridQA. ``2: 12.8\%'' in the chart means that $12.8\%$ instances of training set has exactly $2$ rows with answer-text appearing in them.}
\label{fig:row_dist} \label{fig:mi_stat}
\end{figure}

As depicted in Figure~\ref{fig:row_dist}, for HybridQA, $\sim$40\% of the training instances have the problem of multiple rows containing the correct answer text. In training set of HybridQA dataset, for some instances, the gold answer appears in \emph{19~rows}!

\paragraph{Multi-instance (-row) training:}
A naive way is to label all matches with label $1$ and the rest with label $0$ for training. This reduces the performance of the \textit{RowRetriever} as a large chunk of training data gets incorrect labels.
To address the issue of multiple potentially correct rows we map this problem into a multiple-instance learning setup \citep{mil_97, mi_svm}, with question-row pairs as instances and potential correct rows for a question forming a \textit{bag}.
We are given a question $q$ and table $t$, with row subset $B \subseteq \rows{t}$ labeled~1 (relevant) and the rest, $\rows{t}\setminus B$ labeled~0 (irrelevant).
RowRetriever applied to the retrieval unit of row $r$ is modeled as a function $f(\bm{x}_r)$, where $\bm{x}_r$ is the text constructed in Eqn.~\eqref{eqn:rr_input_format} from row~$r$.
Let $\ell(y_r, f(\bm{x}_r))$ be the binary cross-entropy classification loss, where $y_r\in\{0,1\}$ is the gold label of instance $\bm{x}_i$. For a given table and a question, we define the row retriever loss as
\begin{align}
\min_{r \in B} \ell(1,f(\bm{x}_r)) +
\sum_{r' \notin B} \ell(0,f(\bm{x}_{r'})).
\end{align}
The intuition is that RowRetriever can avoid a loss if it assigns a large score to \emph{any one} of the rows in $B$, whereas it must assign small score to \emph{all} rows not in~$B$.  Apart from this multi-instance loss function, we also deployed a form of curriculum learning~\citep{curriculum}. In early epochs, we only use instances whose labels we are most confident about: negative rows, and questions with only one positive row. In later epochs, we increase the fraction of instances with multiple relevant rows.

\begin{algorithm}[b!]
\begingroup 
\caption{Multi-span AnswerExtractor training.}\label{alg:mat}
\begin{algorithmic}[1]
\Require training instances $D{=}\{(q,t,r_\oplus,\Sigma[r_\oplus])\}$
\State $D_1 \leftarrow \{(q,t,r_\oplus,\Sigma[r_\oplus]){\in}D: \big|\Sigma[r_\oplus]\big|=1\}$
\State $\text{AE}_\text{init} \leftarrow$ train AnswerExtractor on $D_1$ \\
\Comment{initial model based on `easy' cases}
\State $D_{{>}1} \leftarrow \{(q,t,r_\oplus,\Sigma[r_\oplus]){\in}D: \big|\Sigma[r_\oplus]\big|>1\}$
\State $\widehat{D} \leftarrow \varnothing$
\Comment{collects `denoised' instances}
\For{$(q,t,r_\oplus,\Sigma[r_\oplus]) \in D_{{>}1}$}
\State $\sigma^* \leftarrow \argmax_{\sigma\in\Sigma[r_\oplus]} \\
\rule{.5em}{0em}
\text{AnswerExtractor}_{\text{AE1}}(q,\pass{t[r_\oplus,\star]},\sigma)$ \\
\Comment{$\sigma^*$ is the best span among $\Sigma[r_\oplus]$ as per initial model $\text{AE}_\text{init}$}
\State $\widehat{D} \leftarrow \widehat{D} \cup (q,t,r_\oplus,\{\sigma^*\})$
\EndFor
\State $\text{AE}_\text{final} \leftarrow$ train AnswerExtractor on $D_1{\cup}\widehat{D}$
\State\Return{$\text{AE}_\text{final}$} \Comment{refined model}
\end{algorithmic}
\label{alg:multi_ans_train}
\endgroup
\end{algorithm}

\subsection{AnswerExtractor}

In TextQA, answer extraction is solved by a reading comprehension (RC) module \citep{mrcsurvey}. An RC module is usually trained with the query, the passage, and the start and end token positions of the span in the passage where the gold answer is found.
In \shortname, neither start and end index of the span is available (when the answer is a passage span), nor are the table cell coordinates (when the answer is in a table cell). Furthermore, high level supervision of whether the correct answer is a table cell or passage span, is also not available.  This makes the training of AnswerExtractor a challenging task. We tackle this challenge using a multi-span training paradigm.

\paragraph*{Multi-instance (-span) training:}
Recent systems \citep{bert,segal2020simplemultispan} simply consider the first span matching the gold answer text as the correct span and use that for training.  {\itshape This is often an incorrect policy.}  In Figure~\ref{fig:motivating_example}, the correct answer, `2018', occurs multiple times in $\pass{t[r_\oplus,\star]}$, where $r_\oplus$ is the relevant row. There is absolutely no guarantee that the first span in $\pass{t[r_\oplus,\star]}$ matching the gold answer text will be true evidence for answering the question.   Therefore, using the first, or all, matches for training AnswerExtractor can introduce large volumes of training noise and degrade its accuracy.

\begin{algorithm}[b!]
\caption{Joint row+span reranker training.}\label{alg:mir}
\begingroup 
\begin{algorithmic}
\Require Trained RowRetriever and AnswerExtractor;
$K$: number of rows to retain;
$K'$: number of spans to retain;
search space of combining weights~$\mathcal{W}$;
development fold $D = \{(q,t,a)\}$
\For{$w \in \mathcal{W}$}  \Comment{grid search for weights $w$}
\State $\widehat{D} \leftarrow \varnothing$
\For{$(q,t,a) \in D$}
\State $R = \{(r,s)\} \leftarrow$ top-$K$ rows from $\operatorname{RowRetriever}(q,t,K)$ with scores
\For{$(r,s) \in R$}
\State \raggedright $\Sigma=\{(\sigma,s_\text{st},s_\text{en})\}\leftarrow \rule{2em}{0em}\operatorname{AnswerExtractor}(q,t,r,K')$
\State $\vec{s} \leftarrow
\begin{bmatrix} s & s_\text{st} & s_\text{en} \end{bmatrix}$
\State $\operatorname{score}(r,\sigma) \leftarrow w \cdot \vec{s}$
\Comment{combo score}
\EndFor
\State $r_\oplus \leftarrow \argmax_r \operatorname{score}(r,\sigma)$
\State $\widehat{D} \leftarrow \widehat{D} \cup \{(q,t,r_\oplus,a) \}$
\EndFor
\State $\operatorname{perf}(w) \leftarrow$ evaluate AnswerExtractor on~$\widehat{D}$
\EndFor
\State \Return{$\argmax_w \operatorname{perf}(w)$}
\end{algorithmic}
\endgroup
\label{alg:multi_inst_rerank}
\end{algorithm}

Let $\Sigma[r_\oplus]$ be the set of spans in $\pass{t[r_\oplus,\star]}$ that match the gold answer.  Our problem is when $\big| \Sigma[r_\oplus]\big|{>}1$.  Inspired by data programming methods \citep{Ratner2016DataProg}, we propose a multi-span training (MST) paradigm for AnswerExtractor, shown in Algorithm\,\ref{alg:multi_ans_train}. Assuming there is a sufficient number of single-match instances, we train an initial model $\text{AE1}$ on these.  We then use this initial model $\text{AE1}$ to score spans from the noisy instances in $D_{{>}1}$.   Note that this is different from end-task inference, because we are in a highly constrained output space --- we know the answer can only be among the few choices.  The best-scoring span $\sigma^*$ should therefore give us a `denoised' instance. These, combined with the earlier single-span instances, give us a much better training set on which we can train another answer extractor, leading to the final model~$\text{AE2}$.
Appendix\,\ref{app:AnswerExtractor} has more details.

\subsection{RowRetriever feedback (RF)}
\label{subsec:retriever_feedback}

In Algorithm\,\ref{alg:multi_ans_train}, note that a single row $r_\oplus$ is identified in each instance as relevant.  As we have noted before, this is not directly available from training data, because the gold answer may match multiple rows, with no certificate that they are evidence rows.  A trivial approach involves invoking Algorithm\,\ref{alg:multi_ans_train} on all rows containing the gold answer. As expected, this method produced a sub-optimal {AnswerExtractor}.  Instead, we use the trained RowRetriever to identify the most probable row as~$r_\oplus$.


\subsection{Joint row+span reranker (RSR)}
\label{subsec:MRAR}

The final piece in \shortname{} combines the confidence scores of RowRetriever and AnswerExtractor.  Despite the efforts outlined in the preceding sections, they are both imperfect.  E.g., if we retain the top five rows from RowRetriever, gold row recall jumps 8--9\% compared to using only the top one row.  To recover from such situations, we retain the top five rows, along with their relevance scores.  These rows are sent to AnswerExtractor, which outputs its own set of scores for candidate answer spans.  The row+answer reranker implements a \emph{joint selection} across RowRetriever and AnswerExtractor, through a linear combination of their scores, to select the best overall answer.  The weights in the combination are set using a development fold.  These weights can be selected using either grid search or gradient descent, after pinning module outputs.  We do a grid search, shown as Algorithm\,\ref{alg:multi_inst_rerank}.  We shall see that such reranking leads to significant accuracy improvements.

\subsection{Modules for open-domain applications}

\paragraph{TableRetriever:}
For open-domain scenarios where questions are not accompanied by tables, this module retrieves the tables most relevant to a given question.  For this task, we linearize the tables using different special delimiters to distinguish header information, cells and rows. we also prefix the table title in front of the linearized table with a separator. Then we train a dense passage retriever (DPR) \cite{dpr} to give a higher score for a table if it is relevant to the question while computing the dot product of the encoded table and question.  Details about table linearization and DPR training are in Appendix\,\ref{subsec:tab_ret_training}.

\paragraph{RowPassageLinker:}
This module iterates over each row of the tables retrieved by \textit{TableRetriever} and links relevant passages to the row. For every cell in the row, {RowPassageLinker} first searches for nearest neighbour in the passage corpus using a BM25 retriever \cite{chen2017readingdrqa}. Similar to \citet{ottqahen2021open}, RowPassageLinker additionally uses a pre-trained GPT-2 model as context generator for each row and uses the generated context to retrieve more relevant passages from passage corpus.  Details are in Appendix\,\ref{app:RowPassageLinker}.
\begin{table*}
\centering
\adjustbox{max width=\hsize}{  \tabcolsep 5pt
\renewcommand{\arraystretch}{1.0}
\begin{tabular}{l|rrrr|rrrr|rrrr}
\toprule
   & \multicolumn{4}{c|}{\textbf{Table}} & \multicolumn{4}{c|}{{\textbf{Passage}}} & \multicolumn{4}{c}{{\textbf{Total}}}  \\ 
     & \multicolumn{2}{c}{{\textbf{Dev}}} & \multicolumn{2}{c|}{{\textbf{Test}}} & \multicolumn{2}{c}{{\textbf{Dev}}} & \multicolumn{2}{c|}{{\textbf{Test}}} & \multicolumn{2}{c}{{\textbf{Dev}}} & \multicolumn{2}{c}{{\textbf{Test}}}\\ 
  & EM & F1 & EM & F1 & EM & F1 & EM & F1 & EM & F1 & EM & F1  \\ 
\midrule
Table-Only & 14.7 & 19.1 & 14.2 & 18.8 & 2.4 & 4.5 & 2.6 & 4.7 & 8.4 & 12.1 & 8.3 & 11.7 \\
Passage-Only & 9.2 & 13.5 & 8.9 & 13.8 & 26.1 & 32.4 & 25.5 & 32.0 & 19.5 & 25.1  & 19.1 & 25.0\\
HYBRIDER ($\tau=0.8$) \cite{hybridqachen2020} &  54.3 & 61.4 & 56.2 &  63.3 & 39.1 & 45.7 & 37.5 & 44.4 & 44.0 & 50.7 & 43.8 & 50.6  \\
POINTR + MATE $^\dagger$ \cite{eisenschlos2021mate} & \bf 68.6 & \bf 74.2 & 66.9 & 72.3 & 62.8 & 71.9 & 62.8 & 72.9 & 63.4 & 71.0 & 62.8 & 70.2 \\
\midrule
\rowcolor{green!10} \shortname &  68.1 & 73.3 & \bf 68.5 & \bf 74.4 & \bf 66.7 & \bf 75.6 & \bf 64.3 & \bf 73.3 & {\bf 65.5} & {\bf 72.7} & {\bf 64.3} & {\bf 71.9}  \\
\bottomrule
\end{tabular}
}
\caption{End-task performance on dev and test folds of HybridQA, comparing prior systems against \shortname.  $^\dagger$---Systems contemporary to \shortname.}
\label{tab:expt_hybridqa_prior_vs_mitqa}
\end{table*}

\begin{table}[ht]
\centering
\adjustbox{max width=\hsize}{ \tabcolsep 3pt
\begin{tabular}{l|rr|rr}
\toprule
   &  \multicolumn{2}{c|}{{\textbf{Dev}}} & \multicolumn{2}{c}{{\textbf{Test}}}\\ 
   & EM & F1 & EM & F1  \\ 
\midrule
HYBRIDER (Top-1) \cite{hybridqachen2020} & 8.9 & 11.3 & 8.4 & 10.6  \\
HYBRIDER (best Top-K) & 10.3 & 13.0 & 9.7 & 12.8 \\
IR+SBR \cite{ottqahen2021open}& 7.9 & 11.1 & 9.6 & 13.1 \\
FR+SBR \cite{ottqahen2021open}& 13.8 & 17.2 & 13.4 & 16.2 \\
IR+CBR \cite{ottqahen2021open}& 14.4 & 18.5 & 16.9 & 20.9 \\
FR+CBR \cite{ottqahen2021open} & 28.1 & 32.5 & 27.2 & 31.5 \\
CARP$^\dagger$ \cite{zhong2022reasoningcarp} & 33.2 & 38.6 & 32.5 & 38.5 \\
\midrule
\rowcolor{green!10} MITQA & {\bf{40.0}} & {\bf{45.1}} & {\bf 36.4} & {\bf 41.9}\\
\bottomrule
\end{tabular}  }
\caption{End-task performance on dev and test folds of OTT-QA. IR=iterative retriever, FR=fusion retriever. SBR=single block reader, CBR=cross block reader. Best numbers overall are in bold. $^\dagger$---Systems contemporary to MITQA.}
\label{tab:results_v3_2}
\end{table}

\section{Experiments}
\label{sec:Expt}

\subsection{Datasets}

\paragraph{HybridQA} \cite{hybridqachen2020} is the first large scale multi-hop QA dataset that requires reasoning over hybrid contexts of tables and text. It contains 62,682 instances in the train set, 3466 instances in the dev set and 3463 instances in the test set.  HybridQA provides the relevant table and its linked passages with each question, so TableRetriever and RowPassageLinker are not needed.

\paragraph{OTT-QA} \cite{ottqahen2021open} extends HybridQA. It is a large-scale open-domain QA dataset over tables and text which needs table and passage retrieval before question answering. This dataset provides 400K tables and 5M passages as corpus. It has 42K questions in the training set, 2K questions in the dev set, and 2K questions in the test set.

\emph{Multiple rows} containing the answer text pose a major challenge for question answering on these datasets. In HybridQA, $\sim$40\% instances have more than one row in the table matching the answer text exactly. This makes learning to retrieve the most relevant row  nontrivial. 

\emph{Multiple answer spans} pose additional challenges. Further analysis on HybridQA revealed that $\sim$34.5\% instances in the training set have multiple answer spans.
Details are in Appendix \ref{app:expt:datasets}.

\eat{If we take the top row retrieved by the row retriever as the most relevant row then, out of the total 62,682 training instances, answer text can be found in 56,084 instances (89.5\%). Out of these, 36719 (58.5\%) and 19365 (34.5\%) instances have a single span and multiple spans of the answer text respectively. These 19365 instances have a total of 83757 occurrences of answer text. Based on model predictions, answer spans of 7187 (11.46\%) are modified to something other than simply the first occurrence.}

\subsection{Baselines and competing methods}
We compare \shortname's performance with \textbf{HYBRIDER} \cite{hybridqachen2020}, \textbf{CARP} \cite{zhong2022reasoningcarp}, \textbf{MATE} \cite{eisenschlos2021mate} and the methods proposed by \citet{ottqahen2021open}: iterative/fusion retrieveal (\textbf{IR/FR}) + single/cross block reader (\textbf{SBR/CBR}). 
Appendix\,\ref{appendix:baselines} has details.

\subsection{Performance Summary}
\label{subsec:expt:perfsummary}

\paragraph{HybridQA:}
In Table~\ref{tab:expt_hybridqa_prior_vs_mitqa}, we compare the performance of the proposed models on the dev and test sets of HybridQA dataset. We evaluate the performance in terms of exact match (EM) and F1 scores between predicted answer and ground truth answer. We observe that \shortname, 
which incorporates passage filtering, multi instance training and joint row+span reranking achieves the best performance on dev as well as test set in terms of both EM and F1. The final best model achieves $\sim$21\% absolute improvement over HYBRIDER in both EM and F1 on the test splits. At the time of writing, our system also has a $\sim$4\% lead in both EM and F1 over the next best submission on the public leaderboard. Our system  outperforms MATE \cite{eisenschlos2021mate} (a contemporary work reporting performance on HybridQA dataset) by $\sim$1.5--2\%.

\paragraph{OTT-QA:}
In Table\,\ref{tab:results_v3_2}, we compare the performance of the best performing method, \shortname, 
on the dev and test sets of OTT-QA dataset. We report the final answer prediction performance in terms of exact match (EM) and F1 scores. Table \ref{tab:results_v3_2} shows MITQA achieves the best performance on dev as well as test set in terms of both EM and F1. It delivers $\sim$10\% absolute improvement over the best performing baseline by \cite{ottqahen2021open} in both EM and F1 on the test splits. It also achieves $\sim$4\% higher EM on test set when compared to the very recent \eat{arxiv preprint} CARP \cite{zhong2022reasoningcarp}.

\subsection{\shortname{} Ablation Setup}
\label{subsec:comp_abl}

\shortname{} is a complex system with many modules working in concert.  It starts from a base system (RATQA, see below) and then adds several enhancements.  In this section, we compile a list of these enhancements, show their effects on performance, and analyze the results.

\paragraph{RATQA:}
Row retrieval Augmented Table-text Question Answering (RATQA) is a minimal ablation of \shortname.  RATQA includes a \textsc{BERTlarge} \cite{bert} based row retriever trained on standard cross-entropy loss and a \textsc{BERTlarge} based answer extractor.  The answer extractor is trained with all the rows having a string match with the answer text.  During inference, we get the best row from the retriever and apply AnswerExtractor.

\paragraph{MIL:} This is the novel multi instance loss function (Section \ref{subsec:row_retriever}) used to deal with multiple rows getting incorrect labels if they contain the answer text. Without MIL i.e. if a naive cross entropy loss is used, they lead to a noisy training regime.

\paragraph{RF:}
As described in Sec.\,\ref{subsec:retriever_feedback}, we use a pre-trained row retriever to score rows in the train set. This score is used to select the most relevant row while constructing the training data for AnswerExtractor.  For the control case (no RF), we create separate instances for AnswerExtractor from all rows where the gold answer text occurs.

\paragraph{MST:}
Multi-span answer extractor training (Algorithm\,\ref{alg:multi_ans_train}) is used.
For the control case, the leftmost answer span is used.

\paragraph{RSR:}
Algorithm\,\ref{alg:multi_inst_rerank} is used for joint row+span reranking, with $K{=}5$. For the control case, $K{=}1$.

\paragraph{PF:}
PassageFilter (Sec.\,\ref{subsec:PassageFilter} and Appendix\,\ref{app:PassageFilter}) is used to select a limited number of tokens to attach to a linearized row, to fit within the input capacity of BERT.  In the control setting without PF, we concatenate connected passages in left-to-right cell order while constructing the context, and retain the largest prefix accepted by BERT.


\begin{table} \setlength{\fboxsep}{0pt} 
\small\centering
\adjustbox{max width=\linewidth}{%
\begin{tabular}{c|c|c|c|c|rrrr}
\toprule
\multicolumn{5}{c|}{\textbf{Ablations}} & \multicolumn{4}{c}{{\textbf{Total}}}  \\ 
\rotatebox[origin=c]{90}{MIL} &
\rotatebox[origin=c]{90}{RF} &
\rotatebox[origin=c]{90}{MST} &
\rotatebox[origin=c]{90}{RSR} &
\rotatebox[origin=c]{90}{PF} &
\multicolumn{2}{c}{{\textbf{Dev}}} & \multicolumn{2}{c}{{\textbf{Test}}} \\ 
& & & & & EM & F1 & EM & F1  \\ \midrule
\rowcolor{red!8}
 &             &              &              & & 51.6 & 59.5 & 54.0 & 62.1  \\
 & $\checkmark$ &              &              & & 53.5 & 61.2 & 57.3 & 64.6  \\
 & $\checkmark$ & $\checkmark$ &              & & 53.8 & 61.5 & 57.1 & 64.6  \\
 & $\checkmark$ &              & $\checkmark$ & & 58.8 & 66.0 & 59.1 & 66.2  \\
 & $\checkmark$ & $\checkmark$ & $\checkmark$ & & 58.9 & 67.0 & 59.3 & 67.1  \\

 &             &              &              & $\checkmark$ & 60.2 & 68.0 & 57.1 & 65.5 \\
 & $\checkmark$ &              &              & $\checkmark$ & 63.0 & 70.3 & 61.0 & 68.0 \\
 & $\checkmark$ & $\checkmark$ &              & $\checkmark$ & 64.1 & 71.3 & 62.2 & 69.3 \\
 & $\checkmark$ &              & $\checkmark$ & $\checkmark$ & 63.9 & 71.1 & 62.6 & 69.7 \\
 & $\checkmark$ & $\checkmark$ & $\checkmark$ & $\checkmark$ & 64.8 & 71.9 & 63.4 & 70.6 \\

$\checkmark$ &             &              &              & $\checkmark$ &  60.7 & 68.4 & 58.1 & 66.6  \\
$\checkmark$ & $\checkmark$ &              &              & $\checkmark$ &  64.7 & 71.7 & 63.4 & 70.7  \\
$\checkmark$ & $\checkmark$ & $\checkmark$ &              & $\checkmark$ &  64.8 & 71.9 & 63.5 & 70.8  \\
$\checkmark$ & $\checkmark$ &              & $\checkmark$ & $\checkmark$ &  {65.3} & {72.4} & \textbf{64.3} & {71.7}  \\
\rowcolor{green!10}
$\checkmark$ & $\checkmark$ & $\checkmark$ & $\checkmark$ & $\checkmark$ &  {\bf 65.5} & {\bf 72.7} & {\bf 64.3} & {\bf 71.9}  \\

\midrule
\bottomrule
\end{tabular} }
\caption{Ablations of \shortname, starting from the \colorbox{red!8}{RATQA} baseline and progressing to the full \colorbox{green!10}{MITQA} system.}
\label{tab:expt_hybridqa_ablations}
\end{table}

\subsection{\shortname{} Ablation results and analysis}
Table~\ref{tab:expt_hybridqa_ablations} shows the results of ablation experiments.  In the rest of this section, we will discuss the key takeaways. 

\paragraph{Benefits of retrieving row, then span:}
Comparing HYBRIDER in Table\,\ref{tab:expt_hybridqa_prior_vs_mitqa} and RATQA in Table\,\ref{tab:expt_hybridqa_ablations}, we see that our strategy to retrieve correct rows first works better than HYBRIDER, producing $\sim$12\% F1 score improvement even without any other enhancements and without retriever feedback. This shows that identifying the correct/best rows is of utmost importance and brings large benefits.

\paragraph{Multi-Row Training (MIL) benefits:}
Table~\ref{tab:expt_hybridqa_ablations} also gives evidence that training with the new Multi Instance Loss helps RowRetriever increase overall F1 score beyond passage ranking alone.

\paragraph{Multi-Span Training (MST) benefits:}
Multi Span Training (Algorithm \ref{alg:multi_ans_train}) usually boosts performance by 0.5-1\%. This demonstrates the effectiveness of training on denoised data.  

\paragraph{Joint Row+Span Reranking (RSR) benefits:}
Beyond multi-span training (MST) of AnswerExtractor, the joint row+span reranker (RSR) improves F1 score as compared to model variations not applying these strategies. In fact, these enhancements can be applied together --- as seen in Table~\ref{tab:expt_hybridqa_ablations}, model variations with MST+RSR produce the best results.

\paragraph{PassageFilter (PF) benefits:}
While designing PassageFilter, our intent was to minimize the damage from discarded text.  Comparing RATQA+PF against RATQA, we find that not only is PassageFilter effective in this role, but it can, in fact, \emph{increase} F1 score by pruning irrelevant passages before invoking RowRetriever and AnswerExtractor.

\paragraph{Retriever Feedback (RF) benefits:}
In all ablations of \shortname{} that include multi-row training, RF acts as a positive influence, always yielding better F1 scores than ablations without RF.  This translates to better AnswerExtractor performance using less data. With RF, the model is only trained on the best row, while without RF, thrice as much training data is available, but it is more noisy. This also demonstrates the superiority of our row retriever in enhancing answer extractor performance.

\begin{table}[t]
\centering
\adjustbox{max width=\linewidth}{\small
\begin{tabular}{r|c}
\toprule
$K$ & Table retrieval accuracy (\%)\\
\midrule
1  &  41.28\\
5  &  68.15\\
10 &  76.51\\
50 &  88.07\\
\bottomrule
\end{tabular}}
\caption{TableRetriever HITS@$K$, OTT-QA dev set.}
\label{tab:table_retriever}
\end{table}

\subsection{Performance of additional modules}

\paragraph{TableRetriever:}
Given a question, \textit{TableRetriever} retrieves top-k tables from $\sim$400K tables provided in the corpus of OTT-QA. Table~\ref{tab:table_retriever} gives the hit rates at top-$k$ predictions for various values of~$k$.


\paragraph{RowRetriever:}
In Table~\ref{tab:row_selector_res}, we present row retrieval accuracy of our models on the dev split of HybridQA dataset. We also present ablations corresponding to all the modules affecting the accuracy i.e. MIL and PF. 
We observe that passage filtering improves the row retrieval accuracy by $\sim$3\%. Changing standard cross entropy loss to multi instance loss (Section\,\ref{subsec:row_retriever}) further boosts the row retriever accuracy by~$\sim$2\%. 


\begin{table}[t]
\setlength{\fboxsep}{0pt}
\centering
\adjustbox{max width=\linewidth}{\small
\begin{tabular}{c|c|c}
\toprule
\multicolumn{2}{c|}{Ablations} & Row Retrieval \\ 
\rotatebox[origin=c]{90}{MIL} &
\rotatebox[origin=c]{90}{PF} & Accuracy~(\%)  \\ 
\midrule
             &               & 81.39  \\
             &  $\checkmark$  & 84.30  \\
 $\checkmark$ &  $\checkmark$  & 86.38  \\
 \bottomrule
\end{tabular} }
\caption{RowRetriever accuracy, HybridQA dev fold.}
\label{tab:row_selector_res}
\end{table}

\paragraph{PassageFilter:}
We find that average number of tokens in the context for the dev set is 585, with 49\% examples exceeding BERT's maximum token count of 512 (thus needing truncation). We see that, if we follow our passage ranking and filtering strategy before truncation, the answer is retained in the truncated context in around $\sim$1--2\% more dev set examples. Interestingly, the observed performance gain for our answer extractor is slightly larger than this. This can be attributed to the fact that with passage ranking, the correct span more often appears as the first one and gets correctly chosen during back-propagation training for answer extraction.

\eat{After truncation using our passage ranking and filtering strategy, the answer is present in around 1.2\% more examples in the context. Interestingly, the observed performance gain for our answer extractor is slightly larger than this. This is probably because even when the answer is present in both unranked and ranked contexts, the correct span is chosen in the ranked one, as it occurs first in the latter.  \todo{somewhat unclear} }

\subsection{An anecdotal example}
\begin{figure}
\begin{center}
    \includegraphics[width=.9\hsize]{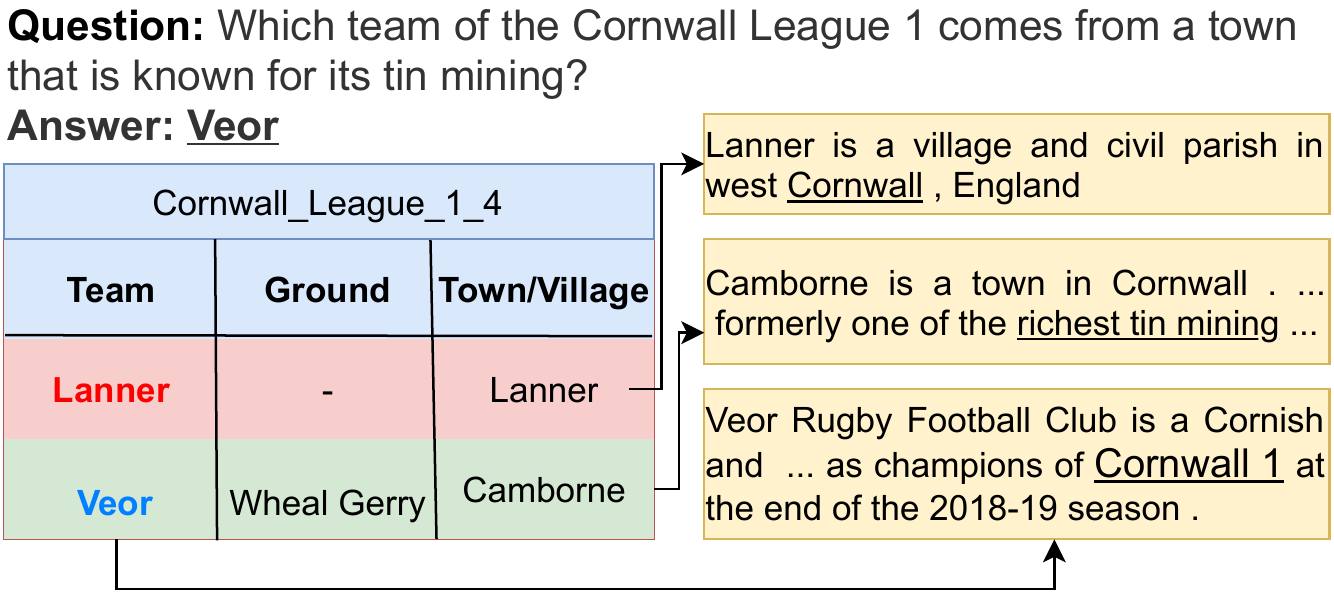}
  \caption{The benefit of MITQA over HYBRIDER.
}
  \label{fig:mitqa_vs_hybrider_2_main}
\end{center}
\end{figure}

Figure\,\ref{fig:mitqa_vs_hybrider_2_main} shows how \shortname{} can outperform at answering questions where the context might cause confusion to both retriever and reader because of multiple matches of important question keywords. 
As shown, HYBRIDER got confused by the presence of `\underline{Cornwall}' in the first row and produced an incorrect answer `{\color{red}Lanner}'. In contrast, \shortname{} predicts the correct answer `{\color{blue}Veor}'.  Appendix\,\ref{app:Anecdotes} shows more examples.

\eat{\subsubsection{Study of critical system params} List 2-3 critical params that govern system performance
Study variation against these.
E.g. beam sizes, number of tokens retained,
number of rows retained.}

\section{Conclusion}
\label{sec:End}

TextTableQA requires reasoning over table cells and linked passage contents. Weak supervision poses a challenge: the target answer might be mentioned in multiple row cells and/or as multiple spans in linked passages. We design a novel QA pipeline that uses multiple row and multiple answer based novel training strategies to identify correct rows first and then use the row cells for relevant passage lookup. We propose efficient strategies for filtering linked passages to retain the most relevant ones for the question, and a novel re-ranker to rank the answers obtained from different rows and their respective linked passages.  
Our system, \shortname, performs better than recent systems on HybridQA and OTT-QA benchmarks, with large improvements in F1 scores. We have also tried different combinations of our proposed strategies to substantiate the benefit from each of them separately.
In future, we would like to explore the following directions:
\begin{enumerate*}[(1)]
\item answering complex numerical questions over hybrid context of table and text, 
\item handling more complex table with structural hierarchies, and
\item enhancing \shortname{} to provide interpretable explanations for answers.
\end{enumerate*}



\section{Limitations}
\label{sec:lim}

Although \shortname{} achieves the best results for TextTableQA benchmarks to date, it still has some limitations, owing to its design, and the type of training data it can access. 

\paragraph{Design policy:}
We have designed \shortname{} as a collection of trainable modules, which are used in a specific sequence. This design has helped us to focus our innovations in specific modules such as multi-row training for RowRetriever, multi-span training for AnswerExtractor, etc., with an eye to boost overall accuracy.  However, the modular design also means that \shortname{} is not fully end-to-end trainable.  Therefore \shortname{} is, in principle, susceptible to compounding error propagation across modules. We view this as an acceptable trade-off while working on HybridQA and OTT-QA, but other data sets may force us to revisit this decision.

\paragraph{Types of queries:} TextTableQA, being a relatively new task, has only two major benchmarks available (HybridQA and OTT-QA), where OTT-QA is an open domain extension of HybridQA. Therefore, the types of queries to which \shortname{} during training are limited to effectively a single large benchmark (HybridQA).  HybridQA --- and consequently OTT-QA --- corpora are similar to Wikipedia articles, not confined to any specific domain.  Further experiments in specific verticals, such as Finance, Retail, and Health are needed to check if \shortname{} affords practical cross-domain adaptation.

Moreover, only a small fraction of queries in HybridQA and OTT-QA need aggregation. Due to their rareness, we have not considered handling aggregation queries through \shortname, which needs additional work in future.

\bibliographystyle{acl_natbib}
\bibliography{aaai22,acl2021}

\clearpage

\appendix

\twocolumn[\begingroup \centering 
\adjustbox{width=\linewidth}{\bfseries \large \ztitle}\\
\bfseries\large (Appendix) \par\smallskip
\endgroup \par \bigskip]

\section{Further details of \shortname{} modules}
\label{app:mitqa_arch_details}

\subsection{TableRetriever and its training}
\label{subsec:tab_ret_training}

In the open domain QA setting (like in OTT-QA) where a designated table $t$ and linked passages $\pass{t[\star,\star]}$ are not provided, we employ the module $\operatorname{TableRetriever}(q)$ to retrieve the most promising tables $T_q \subseteq T$, where $T$ is the corpus of tables.

The training of the TableRetriever module follows the original DPR work \cite{dpr} and its recent application~\cite{glass-etal-2021-robust}, where we first index the linearized tables with \href{https://pypi.org/project/pyserini/}{Anserini}. 
TableRetriever is trained using 
triplet loss over instances of the form $\langle q, t_\oplus, t_\ominus \rangle$,
where $t_\oplus$ is a ground-truth table and
$t_\ominus$ is a \emph{hard negative} \citep{Robinson2021HardNegative} --- an irrelevant table that scores highly with respect to the current scoring model.

To collect hard negative tables $t_\ominus$, we retrieve a pool of tables from a BM25 text retrieval system, and remove the gold table if it is retrieved.  The surviving tables are considered `hard'. To further enhance the robustness of TableRetriever, we select the hard negative table at random from some number of top-scoring hard negative tables. 

The tables and the questions are encoded independently using the same BERT\textsc{base} \cite{bert} model. We later calculate the 
inner product of  question embedding and embedding of all tables to locate the top-scoring relevant tables.



\subsection{RowPassageLinker}
\label{app:RowPassageLinker}

For each table $t{\in}T_q$ returned by TableRetriever and every row $r$ in table~$t$, we use a $\operatorname{RowPassageLinker}(t,r)$ to retrieve the most appropriate passages (from a large corpus of text) and link them to appropriate cells $t[r,c]$.
{RowPassageLinker} first searches for nearest neighbour of the cell text in the passage corpus using a BM25 retriever \cite{chen2017readingdrqa} and retrieves $10$ passages.  Similar to \citet{ottqahen2021open}, RowPassageLinker additionally uses a pre-trained GPT-2 model to generate text from row $t[r,*]$. and uses the generated text as context to retrieve $10$ more relevant passages from the passage corpus.  Specifically, the model takes in the text of $t[r,c]$ as input and outputs additional augmented queries, which are then fed again to the BM25 retriever as queries, to retrieve additional relevant passages. The GPT-2 model is fine-tuned on the supervised pairs of table row (i.e., $t[r,\star]$), header (i.e., $\chdr{t}$) and their hyperlinks  ($t[r,\star] + \chdr{t}$, hyperlink) from in-domain (HybridQA) tables.

\subsection{PassageFilter (PF) and its training}
\label{app:PassageFilter}

Given a question, table, and a set of passages connected to cells in the table, {PassageFilter} ranks the passages based on their relevance to the question. We use Sentence-BERT \cite{reimers-2019-sentence-bert} to get question and passage embeddings and we perform asymmetric semantic search to rank the passages. Asymmetric semantic search is a feature in Sentence-BERT that allows to find a longer passage/document based on a short question.

Passage ranking plays a vital role in row retrieval as well as answer extraction. BERT encoders (used in {RowRetriever} and {AnswerExtractor}) have a limitation that they cannot process sequences of length more than 512 tokens.  Passage ranking ensures that even if we truncate the context to fit BERT, we are unlikely to lose passages most relevant to the question.  Because we do not have supervision about which passages should be ranked higher, we train {PassageFilter} on a similar task of passage ranking given a query on the MS MARCO Passage Retrieval dataset~\cite{bajaj2016ms}. 

Moreover, in case the context contains multiple spans, passage filtering helps to bring the correct answer span at the top, thus reducing the possibility of noisy labels. This is particularly important, because the basic model of answer extractor without multi span training (MST), back-propagates through the first span in the passage matching with the gold answer.

\eat{
\begingroup \color{Tomato2}

\subsection{RowRetriever distant supervision} 
\label{app:RowRetriever}

\todo{decide if this stays}
A row retrieval system that expects supervision in the form of gold retrieval unit (i.e., rows) exacts a high cognitive burden from annotator in preparing training instances. In the case of HybridQA and OTT-QA, we only have final answer-text as supervision, not relevant row/s, cell/s or text span/s. Given a table with connected passages and a question, we identify potential gold rows by exact string matching answer-text on rows (cells and linked texts). We observe that there are multiple rows containing the correct answer-text.

Figure~\ref{fig:row_dist} shows a pie chart for the number of such potentially ``positive" rows that are observed in HybridQA's training set. Over 40\% of the training instances have the problem of multiple rows containing the correct answer text. For some instances, the answer-text appears in as many as 19 rows!
The naive way is to label all matches with label $1$ and the rest with label $0$ for training. This will reduce the performance of {RowRetriever}, as a large chunk of training data will be incorrectly labelled.

\endgroup
}

\subsection{AnswerExtractor text linearization}
\label{app:AnswerExtractor}

A training instance for answer extraction consists of a token sequence generated by concatenating  linearized row contents and passages (linked to cells in the row), together with start and end span indexes of the ground truth answer.  We linearize a row as ``\verb|<column-header>| \verb|is| \verb|<cell-content>|''. This simple linearization bypasses the need to introduce new additional special tokens as column-header and row delimiters, and avoids computationally intensive training of their embeddings.  The concatenated sequential context often exceeds BERT's 512-token limit.  We reduce the probability of the passage containing the ground truth answer getting truncated, by using {PassageFilter} (Appendix\,\ref{app:PassageFilter}).

\section{More Details on Experiments}
\label{appendix:exp}
In this section, we give further details on our experimental approaches.

\subsection{Datasets}
\label{app:expt:datasets}

\textbf{HybridQA}  is the first large scale multi-hop QA dataset that requires reasoning over hybrid contexts of tables and text. It contains 62,682 instances in the train set, 3466 instances in the dev set and 3463 instances in the test set. For the test set, ground truth answers are not available. The authors employ Amazon Mechanical Turk crowd-workers to generate questions based on Wikipedia tables with cells linked to Wikipedia pages.  We split the tables into rows with column headers attached. This enables us to pose the QA problem as row retrieval and answer extraction from the retrieved row.

\textbf{OTT-QA}  extends over HybridQA to make it a large-scale open-domain QA dataset over tables and text which needs table and page retrieval before question answering. This dataset provides 400k tables and 5 million passages as corpus. It has 41,469 questions in the training set, 2,214 questions in the dev set, and 2,158 questions in the test set. According to \cite{ottqahen2021open} , a remarkable difference from original HybridQA \ is that a proportion of questions actually have multiple plausible inference chains in the open-domain setting.


\emph{Multiple rows} containing the answer text pose a major challenge for question answering on these datasets. As depicted in Figure~\ref{fig:row_dist}, for HybridQA, $\sim$40\% instances have more than one row in the table matching the answer text exactly. This makes retrieving the most relevant row highly nontrivial. 

\emph{Multiple answer spans} pose additional challenges. Further analysis on HybridQA revealed that $\sim$34.5\% instances in the training set have multiple answer spans.

\subsection{Baselines and Competing Methods}
\label{appendix:baselines}
\paragraph{\bfseries HYBRIDER}
We compare our model's performance with the standard HYBRIDER \citep{hybridqachen2020} baseline. HYBRIDER uses a two phase process of linking and reasoning to answer questions over heterogeneous context of table and text. This approach attempts to use cell as a unit for linking, hopping and answer prediction.

\paragraph{\bfseries Iterative and Block Retrieval}
These models are proposed by \citet{ottqahen2021open} and are combinations of Iterative/Fusion retrievers and Single/Cross readers. Fusion retrieval uses  “early fusion" strategy to group tables and passages as fused blocks before retrieval. Single Block Reader feeds top-k blocks independently to the reader and selecting the best answer. Cross Block Reader concatenates top-k blocks together to the reader, and generates a single joint answer string.

\paragraph{\bfseries MATE}
MATE \cite{eisenschlos2021mate} models the structure of large Web tables. It uses sparse attention in a way that allows heads to efficiently attend to either rows or columns in a table. To apply it on HybridQA, the authors propose $PointR$, which expands a cell using description of its enitities, selects an appropriate expanded cell and then reads the answer from it.

\paragraph{\bfseries CARP} CARP \citep{zhong2022reasoningcarp} is a chain-centric reasoning and pre-training framework for table-and-text question answering. It
first extracts explicit hybrid chain to reveal the intermediate reasoning process leading to the answer across table and text. The hybrid chain then provides a
guidance for QA, and explanation of the intermediate reasoning process.

\paragraph{\bfseries Other baselines}
These can be found on the respective challenge leaderboards.\footnote{HybridQA: \url{https://competitions.codalab.org/competitions/24420} \\ OTT-QA: \url{https://competitions.codalab.org/competitions/27324} } There are no linked papers to the submissions as yet. We compare \shortname's test performance against all of them.
\subsection{Implementation Details}
MITQA is implemented using Pytorch version 1.8 and Huggingface's transformers\footnote{\url{https://huggingface.co/}} \cite{wolf-etal-2020-transformers} library.  We train our models using two NVIDIA A100 GPUs. We train the row retriever and answer extractor for 5 epochs and select the best model based on dev fold performance. We optimize the model parameters using AdamW algorithm with a learning rate of $5{\times}10^{-5}$ and a batch size of~24.  We set per-GPU train batch size to 16 while training the answer extractor.  We evaluate final answers using EM (exact match) and F1 metrics. 

\textbf{Average Runtime}: Overall training of MITQA takes approximately 24 hours on A100 gpu.

\textbf{Hyperparameter Details:} We tune hyper-parameters based on loss on validation set. We use the following range of values for selecting the best hyper-parameter \\ 
$\bullet$ \textbf{Batch Size:} 8, 16, 32 \\
$\bullet$ \textbf{Learning Rate:} 1e-3, 1e-4, 1e-5, 1e-6, 3e-3, 3e-4, 3e-5, 3e-6, 5e-3, 5e-4, 5e-5, 5e-6 

\begin{figure}[t]
\begin{center}
    \includegraphics[scale=0.62]{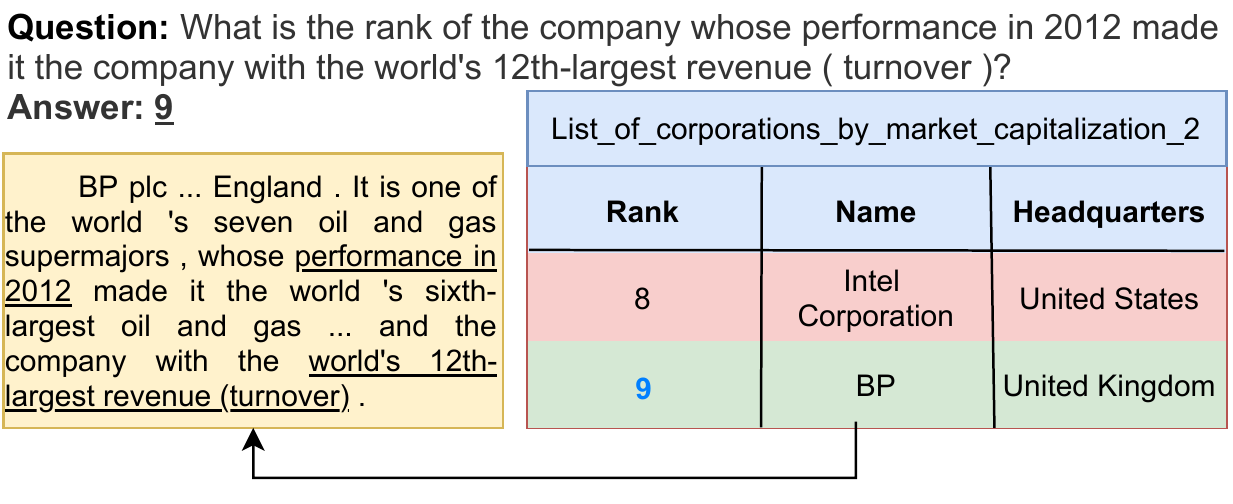}
  \caption{\shortname{} is able to extract answer even if the answer is only present in the table as a cell value. The correct answer is highlighted in {\color{blue} blue}. Despite having other numbers in the table and phrases mentioning ranks like `seven', `sixth-largest', etc. in the passage MITQA was able to predict the correct answer from the table.}
  \label{fig:ans_in_table}
\end{center}
\end{figure}

\section{Anecdotes of Gains}
\label{app:Anecdotes}

\subsection{Answer in Table Cell}
We present in Figure \ref{fig:ans_in_table} an example where MITQA is able to predict the answer correctly even when the correct answer is in a table cell and not a span in the passages.

\subsection{Benefits of Multi Span Training (MST)}
Figure~\ref{fig:mat_eg} shows an example where MST leads the model to train on the correct answer span, thereby leading to a less noisy training regime.

\begin{figure}[ht]
\noindent\fbox{\begin{minipage}{.97\hsize}
\raggedright\scriptsize\sffamily
{\bfseries Question:} What was the mascot of the college of Ryan Quigley ? \\
{\bfseries Answer:} Eagles \\
{\bfseries Context:} Original NFL team is Chicago Bears . Player is \underline{Ryan Quigley} . Pos is P . \underline{College is Boston College} . Conf is ACC . Ryan Andrew Quigley ( born January 26 , 1990 ) is an American football punter who is currently a free agent . He was signed by the Chicago Bears after going undrafted in the 2012 NFL Draft . He played college football at Boston College . He has played for the New York Jets , Philadelphia \textbf{\textcolor{red}{Eagles(7.73)}} , Jacksonville Jaguars , Arizona Cardinals and Minnesota Vikings . The 2011 \underline{Boston College} \textbf{\textcolor{blue}{Eagles(0.03)}} football team represented Boston College in the 2011 NCAA Division I FBS football season . The \textbf{\textcolor{red}{Eagles(6.27)}} were led by third year head coach Frank Spaziani and played their home games at Alumni Stadium . ...
\end{minipage}}
\caption{Benefits from MAT. The model loss is shown in brackets along with the spans. It is clear that the correct mention (in {\textcolor{blue}{blue}}) rightly gets the lowest loss while the ones which are irrelevant (in {\textcolor{red}{red}}) have higher losses. Contexts that can potentially help answer the question are underlined. The first `Eagles' in entirely irrelevant as it refers to a different team. The second one is the best answer by far. The third occurrence refers to the correct team, but lacks as good a context as the second (for model learning).}
\label{fig:mat_eg}
\end{figure}

\subsection{Benefits of Row Span Re-ranker (RSR)}
\begin{figure}[t]
\centering\includegraphics[width=.9\hsize]{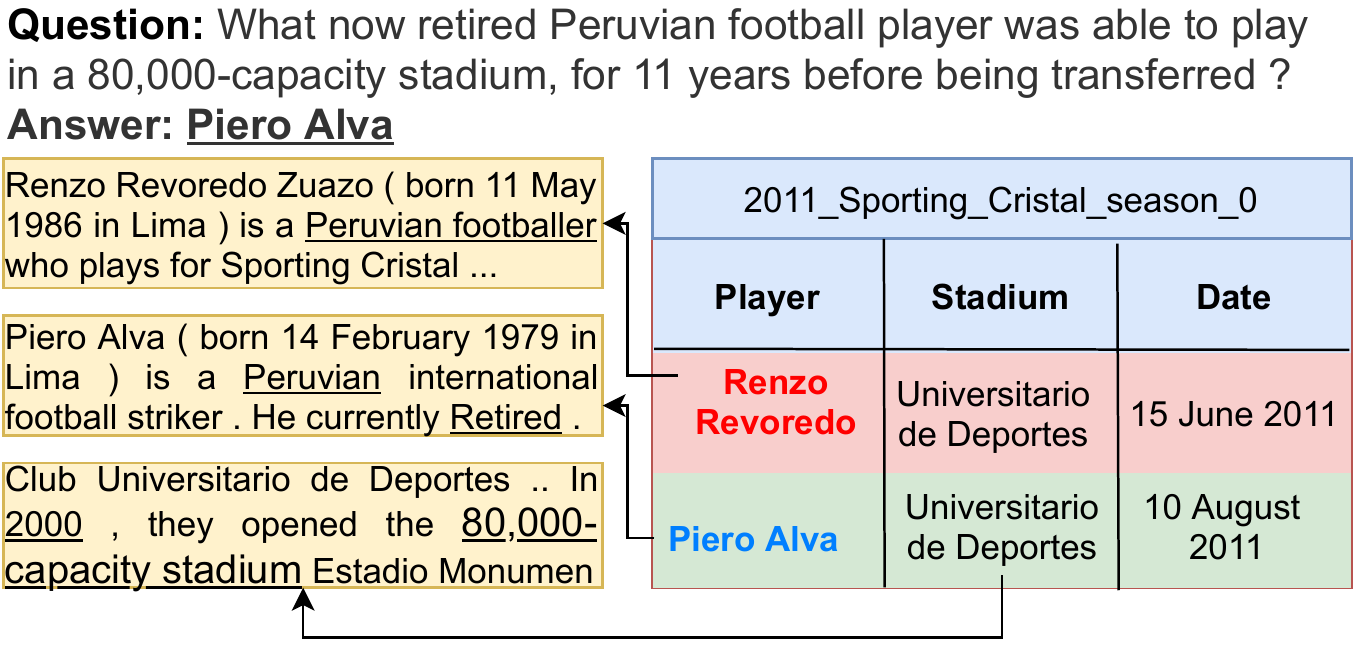}
  \caption{The benefit of Row Span Re-ranker. The correct answer is highlighted in {\color{blue}blue} and the incorrect answer is highlighted in {\color{red}red}. Both `Piero Alva' and `Renzo Revoredo' played in the same stadium (`Universitario de Deportes'). But only `Piero Alva' (answer after reranking) has retired while `Renzo Revoredo' (answer before reranking) has not. Thus, the RSR helps rank the correct answers higher than the incorrect ones in similar contexts and confusing scenarios.}
  \label{fig:MRR_eg}
\end{figure}

Figure \ref{fig:MRR_eg} depicts an instance where RSR is able to rectify the error made by MITQA. The incorrect answer also appeared in a context very similar to the context of correct answer but Multi-Row Re-ranker is able to rank the correct answer higher than the incorrect answer. \eat{Also, in Figure \ref{fig:mitqa_vs_hybrider}, RSR ranks the correct answer higher even though both the answers seems plausible answers.}

\makeatletter
\setlength{\@fptop}{0pt}
\makeatother

\end{document}